\documentclass[sigconf]{acmart}

\setcopyright{none}
\renewcommand\footnotetextcopyrightpermission[1]{} 
\usepackage{graphicx, array, blindtext}
\usepackage{changepage}
\usepackage{subfigure}
\usepackage{alphalph}
\usepackage{float}

\usepackage{array}
\newcolumntype{P}[1]{>{\centering\arraybackslash}p{#1}}
\usepackage{algorithm}
\usepackage{algpseudocode}

\AtBeginDocument{%
  \providecommand\BibTeX{{%
    \normalfont B\kern-0.5em{\scshape i\kern-0.25em b}\kern-0.8em\TeX}}}

\begin{document}

\title{CMA-CLIP: Cross-Modality Attention CLIP for Image-Text Classification}


\author{Huidong Liu}
\authornote{All three authors contributed equally to this research.}
\email{huidliu@cs.stonybrook.edu}
\orcid{1234-5678-9012}
\affiliation{%
  \institution{Stony Brook University}
  \streetaddress{100 Nicolls Rd}
  \city{Stony Brook}
  \state{NY}
  \country{USA}
  \postcode{11794}
}
\author{Shaoyuan Xu}
\authornotemark[1]
\email{shaoyux@amazon.com}
\affiliation{%
  \institution{Amazon Inc.}
  \streetaddress{321 Terry Ave N}
  \city{Seattle}
  \state{WA}
  \country{USA}
  \postcode{98109}
}

\author{Jinmiao Fu}
\authornotemark[1]
\email{jinmiaof@amazon.com}
\affiliation{%
  \institution{Amazon Inc.}
  \streetaddress{321 Terry Ave N}
  \city{Seattle}
  \state{WA}
  \country{USA}
  \postcode{98109}
}

\author{Yang Liu}
\email{yliuu@amazon.com}
\author{Ning Xie}
\email{xining@amazon.com}
\author{Chien-Chih Wang}
\email{ccwang@amazon.com}
\affiliation{%
  \institution{Amazon Inc.}
  \streetaddress{321 Terry Ave N}
  \city{Seattle}
  \state{WA}
  \country{USA}
  \postcode{98109}
  }

\author{Bryan Wang}
\email{brywan@amazon.com}
\author{Yi Sun}
\email{yisun@amazon.com}
\affiliation{%
  \institution{Amazon Inc.}
  \streetaddress{321 Terry Ave N}
  \city{Seattle}
  \state{WA}
  \country{USA}
  \postcode{98109}
  }



\begin{abstract}
Modern Web systems such as social media and e-commerce contain rich contents expressed in images and text. Leveraging information from multi-modalities can improve the performance of machine learning tasks such as classification and recommendation. In this paper, we propose the Cross-Modality Attention Contrastive Language-Image Pre-training (CMA-CLIP), a new framework which unifies two types of cross-modality attentions, sequence-wise attention and modality-wise attention, to effectively fuse information from image and text pairs. The sequence-wise attention enables the framework to capture the fine-grained relationship between image patches and text tokens, while the modality-wise attention weighs each modality by its relevance to the downstream tasks. In addition, by adding task specific modality-wise attentions and multilayer perceptrons, our proposed framework is capable of performing multi-task classification with multi-modalities. 

We conduct experiments on a Major Retail Website Product Attribute (MRWPA) dataset and two public datasets, Food101 and Fashion-Gen. The results show that CMA-CLIP outperforms the pre-trained and fine-tuned CLIP by an average of 11.9\% in recall at the same level of precision on the MRWPA dataset for multi-task classification. It also surpasses the state-of-the-art method on Fashion-Gen Dataset by 5.5\% in accuracy and achieves competitive performance on Food101 Dataset. Through detailed ablation studies, we further demonstrate the effectiveness of both cross-modality attention modules and our method's robustness against noise in image and text inputs, which is a common challenge in practice. 
\end{abstract}




\maketitle
\pagestyle{plain}

\section{Introduction}
Inspired by the recent rise of the pre-trained NLP models such as BERT \cite{devlin2018bert}, learning to classify image-text pairs for vision-language ($VL$) tasks using Transformer \cite{NIPS2017_3f5ee243} based encoders has received much attention as both modalities can be informative and beneficial to each other. Current methods can be classified into two main categories: one-stream methods and two-stream methods. One-stream methods capture the cross-modality attention across image and text by concatenating them at early stage and input the concatenated feature into one unified transformer encoder. Two-stream methods first extract the image and text features using two separate encoders and then learn their cross-modal relationship through various methods such as contrastive learning \cite{chen2020simple}, etc.

Among those one-stream and two-stream methods, the Contrastive Language-Image Pre-Training (CLIP) \cite{radford2021learning} has achieved great success recently. CLIP is trained on the WebImageText (WIT) Dataset which consists of 400 million image-text pairs collected from a variety of publicly available sources on the Web and achieves many state-of-the-art results in zero-shot learning tasks, pre-training tasks, and supervised classification tasks when a linear probe is added on top of it.

\begin{figure*}[h]
\includegraphics[width=7in]{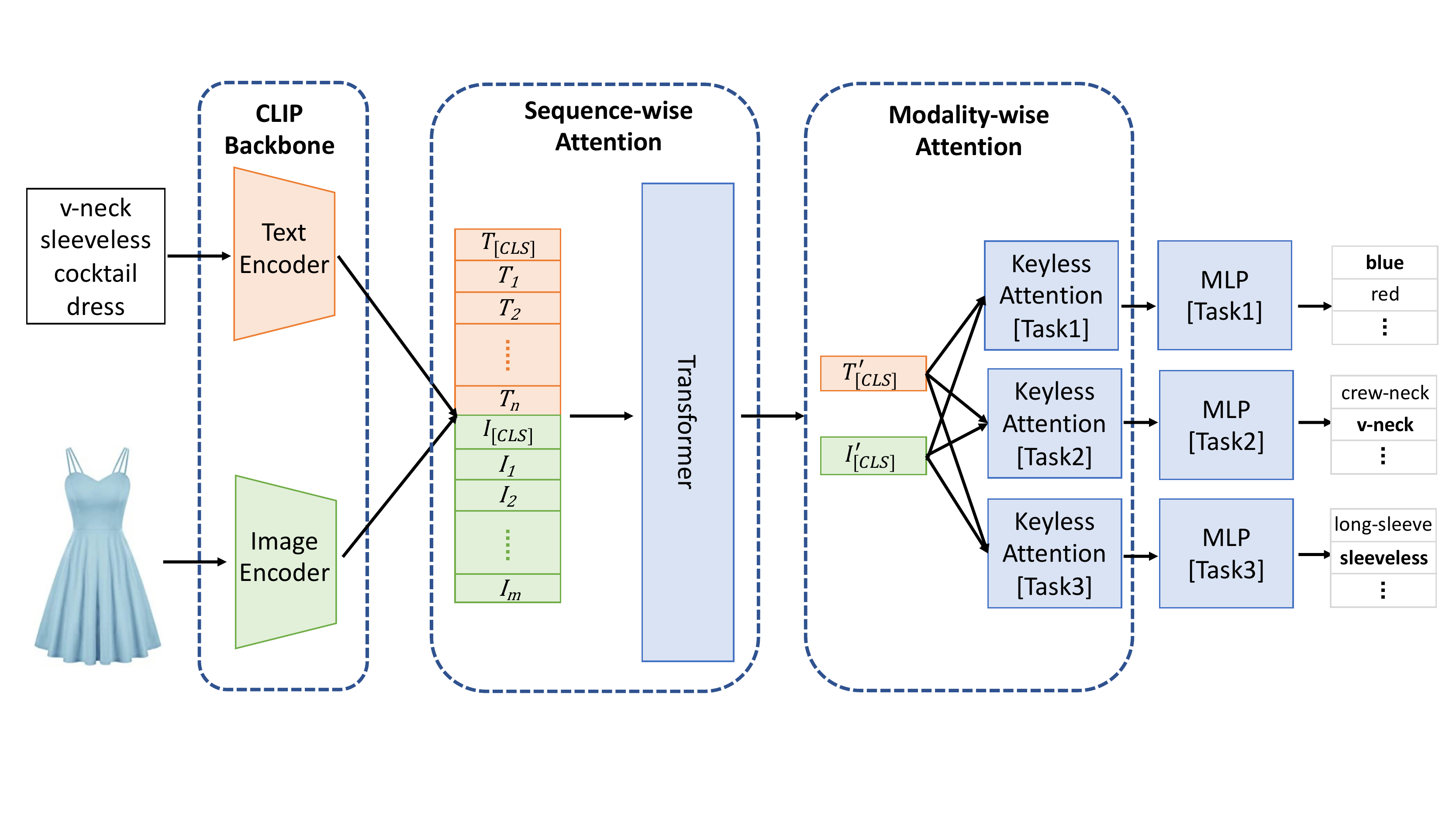}
\vspace{-50pt}
\caption{The pipeline of our proposed CMA-CLIP. }
\label{fig: cma_pipeline}
\end{figure*}

Despite of CLIP's \cite{radford2021learning} strength, it is mainly designed for zero-shot image classification, resulting in the limitation of its ability to leverage both image and text input when available. Since the training of CLIP only involves global image and text features, thus the fine-grained relationship between image patches and text tokens are not modeled. Such relationship is useful in fine-trained classification tasks, especially in the situations where only a small proportion of image patches or text tokens are related to the classification tasks. Moreover, since it chooses the user-defined textual description called "prompts" with class value that matches the most with the image as the classification result, significant efforts to engineer the prompts for optimizing downstream tasks are required. Last but not least, in practice, it is quite common for the image-text pairs to contain noise. For instance, on E-commerce websites, some images or text could be irrelevant to the product due to catalog errors. In social media apps, users might enter irrelevant textual comments or upload unrelated images. Treating input from both modalities equally in such situation may lead to poor classification performances as one of the modalities could be pure noise. To address the aforementioned issues, in this paper we propose the Cross-Modality Attention CLIP (CMA-CLIP). Our contributions include:

\begin{itemize}
  \item We combine CLIP with a sequence-wise attention module, which refines the CLIP-generated image and text embeddings by modeling the relationship among the embedding of the sequence of image patches and text tokens. This transformer-based module makes the embedding more context-aware. $E.g.$, the embedding of the black image patches are more correlated with the `$black$' token in text. We experimentally prove that such refinement can improve the performance of classification tasks.  
  \item We adopt a modality-wise attention module to assign learnable weights to each modality that measures its relevance to the classification task. The impact of the irrelevant modality will be dampened, and therefore our network is robust against noisy image or text inputs, which is a common challenge in practice.
  \item We add task specific modality-wise attentions and MLP heads on top of the sequence-wise attention module, so that the same network can be leveraged for multi-task classification. Moreover, compared with CLIP, this architecture enables the network to leverage both image and text inputs in both training and inference stage.
  \item On the MRWPA, CMA-CLIP outperforms the raw CLIP and the fine-tuned CLIP (fine-tuned using image-text pairs from a major retail website) on the classification of three product attributes by 10.9\% and 12.9\% in recall at the same level of precision. It also improves the state-of-the-art performance \cite{zhuge2021kaleido} on Fashion-Gen Dataset from 88.1\% to 93.6\% in accuracy and achieves competitive performance on Food101 Dataset against \cite{kiela2019supervised}.
\end{itemize}

\section{Related Work}

Current multi-modality learning methods are mainly one-stream and two-steam where one-stream methods use a single Transformer encoder to process the concatenated image and text embedding, while two-stream methods use both image encoder and text encoder to extract image and text embeddings at early stage and then learn their cross-modal relationship.

The one-stream methods, such as ViLBERT \cite{lu2019vilbert}, VisualBERT \cite{li2019visualbert}, VL-BERT \cite{su2019vl}, Unicoder-VL \cite{li2020unicoder}, ImageBERT \cite{qi2020imagebert} and Unified VLP \cite{zhou2020unified}, concatenate the image's Region-Of-Interest (ROI) patches and text tokens as the input tokens for BERT \cite{devlin2018bert}. These models are typically pre-trained using tasks including Masked Language Modeling (MLM), Masked Region Modeling (MRM), Multi-Model Alignment Prediction (MMAP). The UNITER \cite{chen2020uniter} and OSCAR \cite{li2020oscar} incorporate additional pre-training tasks. The UNITER uses the Optimal Transport (OT) \cite{villani2009optimal} to model the relationship between the image patches and the text tokens. The OSCAR uses the object categories detected by the Faster-RCNN \cite{ren2015faster} and encodes the category text as additional input tokens to BERT. Instead of using the Faster-RCNN to detect ROIs, methods such as ICMLM \cite{sariyildiz2020learning}, Pixel-BERT \cite{DBLP:journals/corr/abs-2004-00849}, and SOHO \cite{huang2021seeing} use a CNN to extract the feature maps of an image and use the depth vectors in feature maps as image tokens. Such configuration is able to capture the semantic connection between image pixels and text tokens, which is overlooked by region based image features extracted by Faster-RCNN. Similar to ICMLM, the VirTex \cite{desai2021virtex} also uses the depth vectors in feature maps as image tokens and input the image and text tokens into a forward transformer decoder and a backward transformer decoder. Instead of feeding the whole image or ROI into a CNN, methods like FashionBERT \cite{gao2020fashionbert}, KaleidoBERT \cite{zhuge2021kaleido}, and ViLT \cite{pmlr-v139-kim21k} cut an image into patches and treat each patch as an "image token". For FashionBERT, it uses a pre-trained image model such as InceptionV3 \cite{7780677} or ResNeXt-101 \cite{pub.1095850372} to extract image features. Different from FashionBERT, KaleidoBERT adopts the SAT \cite{pmlr-v37-xuc15} network to generate description of salient image patches aiming to find an approximate correlation between image patches and text tokens to serve their pre-training tasks, $i.e.$, Aligned Masked Language Modeling (AMLM), Image and Text Matching (ITM), and Aligned Kaleido Patch Modeling (AKPM). ViLT differs from all above-mentioned methods by simply applying linear projection on flattened image patches which greatly reduce the model size, thus leading to significant runtime and parameter efficiency.

\begin{table*}[t]
\centering
\begin{tabular}{ || m{0.15\linewidth}  m{0.23\linewidth} m{0.15\linewidth}  m{0.35\linewidth} ||} 
\hline
Dataset & Image & Label & Text \\  
\hline\hline
MRWPA &\includegraphics[scale=0.15]{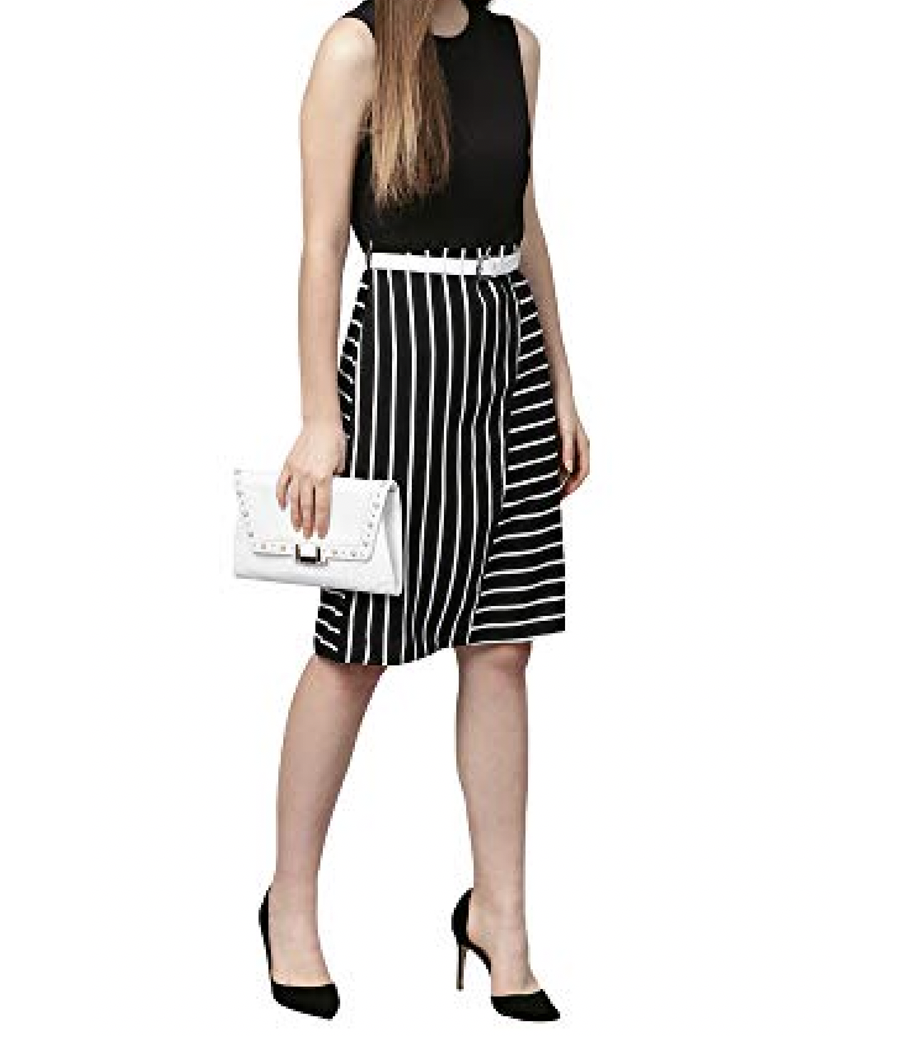}&Striped&Women Multi Striped Sheath Dress\\
\hline
Food101 Dataset&\includegraphics[scale=0.13]{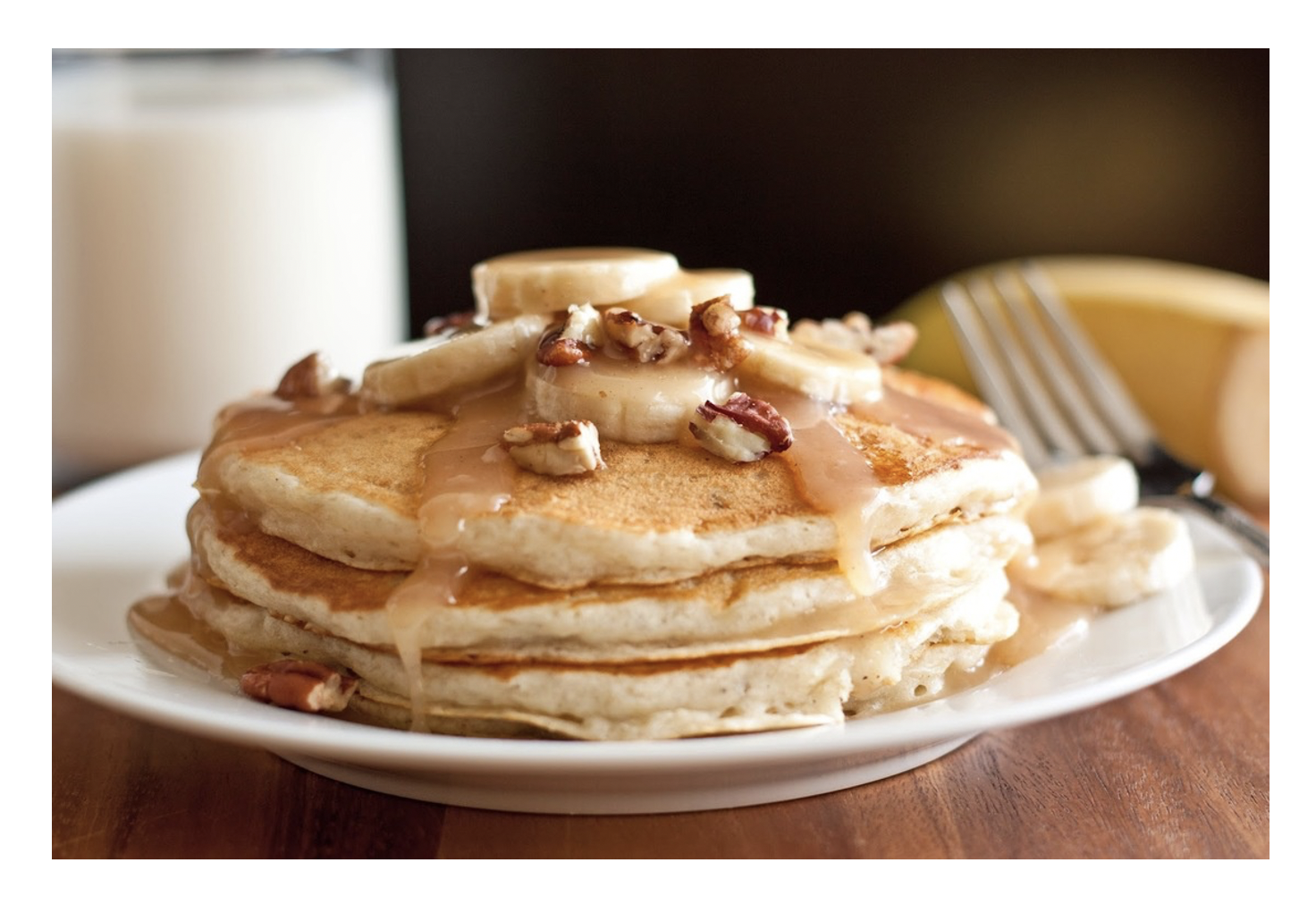}&Pancakes&Banana Bread Pancakes with Cinnamon Cream Cheese Syrup - Cooking Classy\\
\hline
Fashion-Gen Dataset&\includegraphics[scale=0.25]{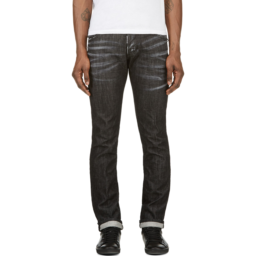}&Jeans&Slim-fit jeans in dark blue. Distressing throughout. Fading at front. Textured black leather logo patch at back waist. Silver-tone metal logo plaque at back pocket. Contrast stitching in tan. Red logo tab at button-fly.\\
\hline
\end{tabular}
\bigskip
\caption{Data example for each of the datasets}
\label{tb: Datasets}
\end{table*}

\begin{algorithm*}[tb]
   \caption{Training Process of CMA-CLIP. Line 4-16: Warm-up Stage, Line 18-21: End-to-End Training Stage, Line 23-26: Tuning Stage}
   \label{alg:cma-clip}
\begin{algorithmic}[1]
 \Require Image-text pairs and their labels of K tasks $(\mathcal{X}, \mathcal{Y}, \mathcal{L}_{1}, \mathcal{L}_{2}, ..., \mathcal{L}_{K})$.
   \State $CLIP$ denotes the CLIP model including the image and text encoders. $CMA_{SA}$ denotes the sequence-wise attention model. $CMA_{MA}$ denotes the modality-wise attention model. $MLP$ denotes the multi-layer perception head for classification.
   \State \textbf{Training:}
   \State Use the image encoder and the text encoder from CLIP, and freeze their weights.
   \State Set $lr_{CLIP} = 0$, $lr_{CMA} = 1e-5$, $lr_{MLP} = 1e-5$ \Comment{since the CLIP module is frozen, we set $lr_{CLIP}$ to be 0}
   \While {$i < MaxEpoch1$}
   \State $(\hat{\mathcal{X}}_{CLIP}, \hat{\mathcal{Y}}_{CLIP})$ = $CLIP(\mathcal{X}, \mathcal{Y})$ \Comment{apply CLIP's image and text encoders on $\mathcal{X}$ and $\mathcal{Y}$}
   \State $\hat{\mathcal{X}}\hat{\mathcal{Y}}_{concat}$ = $concat(\hat{\mathcal{X}}_{CLIP}, \hat{\mathcal{Y}}_{CLIP})$ \Comment{concatenate the image feature $\hat{\mathcal{X}}_{CLIP}$ and the text feature $\hat{\mathcal{Y}}_{CLIP}$} 
   \State $(\hat{\mathcal{X}}_{SA}, \hat{\mathcal{Y}}_{SA})$ = $CMA_{SA}(\hat{\mathcal{X}}\hat{\mathcal{Y}}_{concat})$ \Comment{apply the sequence-wise attention module to the concatenated feature $\hat{\mathcal{X}}\hat{\mathcal{Y}}_{concat}$}
   \For {k = 1 to K}
   \State $\mathcal{U}^{k}_{MA}$ = $CMA^{k}_{MA}(\hat{\mathcal{X}}_{SA}, \hat{\mathcal{Y}}_{SA})$ \Comment{apply the modality-wise attention module to get the weighted average feature $\mathcal{U}^{k}_{MA}$}
   \State $\hat{\mathcal{L}}_{k}$ = $MLP(\mathcal{U}^{k}_{MA})$ \Comment{use the MLP head for classification task to get prediction $\hat{\mathcal{L}}_{k}$}
   \State $Loss_{k}$ = $Softmax(\mathcal{L}_{k}, \hat{\mathcal{L}}_{k})$ \Comment{compute Softmax Loss between ground truth $\mathcal{L}_{k}$ and prediction $\hat{\mathcal{L}}_{k}$}
   \EndFor
   \State $Loss$ = $Loss_{1}$ + $Loss_{2}$ + ... + $Loss_{K}$
   \State $\theta^{i+1}$ = $Adam(Loss, \theta^{i})$ \Comment{update model parameters using Adam}
   \EndWhile
   \State Release all modules. Fine-tune on the best check-points from the previous stage.
   \State Set $lr_{CLIP} = 1e-5$, $lr_{CMA} = 1e-5$, $lr_{MLP} = 1e-5$ \Comment{since all modules are released, all learning rates are set to $1e-5$}
   \While {$i < MaxEpoch2$}
   \State Repeat Line 6-15.
   \EndWhile
   \State Freeze the image encoder, the text encoder, the sequence-wise attention transformer. Fine-tune on the best check-points from the previous stage.
   \State Set $lr_{CLIP} = 0$, $lr_{CMA} = 0$, $lr_{MLP} = 1e-5$ \Comment{since both CLIP and CMA modules are frozen, we set $lr_{CLIP}$ and $lr_{CMA}$ to be 0}
   \While {$i < MaxEpoch3$} 
   \State Repeat Line 6-15.
   \EndWhile
\Ensure $CLIP$ model, $CMA_{SA}$ model, $CMA_{MA}$ model, and $MLP$ model.
\end{algorithmic}
\end{algorithm*}

The two-stream methods are mainly motivated by self-supervised learning methods \cite{chen2020simple, he2020momentum, grill2020bootstrap, chen2021exploring, chen2020improved, chen2021empirical, caron2020unsupervised, chen2020big}. In self-supervised learning, two views ($e.g.$, two augmentations), of a single image are forwarded into one network respectively. Their outputs are compared using the contrastive loss \cite{chen2020simple}, so that the two views of the same image are much similar than the two views from two different images. The ConVIRT \cite{zhang2020contrastive} adopts this idea on the self-supervised learning of image-text pairs. Two networks are used to extract the image and text features respectively. The image and text features from the paired image-text input is trained to be much similar than the unpaired ones. CLIP \cite{radford2021learning} is a simplified version of ConVIRT, where the text in each image-text pair is a single sentence instead of a pool of sentences as in ConVIRT. Similar as CLIP, BriVL \cite{huo2021wenlan} uses MoCo \cite{he2020momentum} which is a more advanced cross-modal contrastive learning algorithm to help train the network with limited GPU memory by leveraging more negative samples. The ALIGN \cite{jia2021scaling} collects 1.8 billion image-text pairs, and adopts a similar network architecture as CLIP. The performance of ALIGN is comparable to CLIP on the ImageNet dataset for the classification task, Flickr30K and MSCOCO datasets for image-text retrieval task. 

In order to perform image-text classification tasks, a classification layer needs to be added on top of the image-text embeddings of the pre-trained models such as VL-BERT \cite{su2019vl}, UNITER \cite{chen2020uniter}, et al. The MMBT \cite{kiela2019supervised} is specifically designed for image-text classification tasks. Unlike VL-BERT or UNITER, MMBT directly loads weights from BERT which does not require pre-training. In MMBT, ResNet \cite{he2016deep} is used to extract image features. The image features are projected into the same space as the text tokens, used as image tokens and feed into a BERT together with text tokens. A linear layer is added on the classification embedding to perform supervised tasks. 

However, both one-stream and two-stream methods have their own inadequacies. One-stream methods heavily rely on pre-trained Faster-RCNN \cite{ren2015faster} or ResNet \cite{he2016deep} to extract image feature which does not support end-to-end training of the whole network, and therefore, the extracted image and text features are not optimized to model the image-text relationship. Two-stream methods only focus on learning global image and text features, and cannot capture the fine-grained relationship between image patches and text tokens.

In order to overcome the aforementioned disadvantages, our proposed method fuses both one-stream and two-stream architectures which complements each other's inadequacies. It leverages the pre-trained CLIP, a two-stream architecture, to capture the overall alignment between image and text. Subsequently, we add a sequence-wise attention module, which is a transformer based cross-modality attention module used in most one-stream architectures, to capture the fine-grained relationship between image patches and text tokens. SemVLP \cite{Li2021SemVLPVP} applies similar fusion logic. The difference between our method and SemVLP is that, SemVLP leverages the same cross-modality attention module to capture both high-level and fine-grained relationship between image and text. It was pre-trained on tasks such as MLM and MRM, whereas CLIP was directly pre-trained to maximize the overall image-text alignment through contrastive learning. More importantly, SemVLP does not consider the situation where one of the modalities is irrelevant to the downstream classification tasks due to input noises, which is a common challenge in practice. To handle such situation, we add a modality-wise attention module, which learns the importance of both modalities so that the irrelevant modality can be dampened for the classification tasks. At last, by adding task specific modality-wise attentions and MLPs, our model is able to perform multi-task classifications.

\section{Method}
The rest of the paper is arranged as follows: In Section 3, we first give a brief review of CLIP, and then we introduce our proposed CMA-CLIP with detailed explanation of each component. In Section 4, we introduce the datasets that we use, the corresponding experimental results, the visualization of the sequence-wise attention module, and the ablation study to prove the effectiveness of modality- and sequence-wise attention modules. In Section 5, we conclude this paper and elaborate our future work.

\subsection{Contrastive Language-Image Pre-Training (CLIP)}
The Contrastive Language-Image Pre-Training (CLIP) consists of an image encoder and a text encoder.  For each image-text pair, the image and text encoders project the pair into an image and text embedding in the same multi-modal space. Given $N$ image-text pairs, the training objective of CLIP is to maximize the cosine similarity of the paired image and text embedding while minimize the cosine similarity of the unpaired ones.  

During inference, for a classification task with $K$ classes, it first uses the $K$ class values to construct $K$ prompts such as `A photo of \textbf{\{class value\}}'. These $K$ prompts are then projected to $K$ text embeddings by the text encoder. For any given image, it is projected to an image embedding by the image encoder, then CLIP computes the cosine similarities between the image embedding and those $K$ text embeddings. The class value with the largest similarity is then considered as the class prediction. 

CLIP is trained using WIT Dataset which contains 400 million image-text pairs collected from the Web. According to the results reported in \cite{radford2021learning}, its zero-shot classification performance surpasses the supervised linear classifier fitted on ResNet50 \cite{he2016deep} features on datasets such as StanfordCars \cite{krause20133d}, Country211 \cite{radford2021learning}, Food101 \cite{bossard14}, and UCF101 \cite{soomro2012ucf101} etc.

\subsection{The Cross-Modality Attention CLIP (CMA-CLIP)}

CLIP focuses on the learning of the global image and text features. In CMA-CLIP, we build a sequence-wise attention module to capture the fine-grained relationship between the image patches and the text tokens such as the black image patches and the `black' tokens in text. This module leverages the transformer architecture \cite{NIPS2017_3f5ee243}. It takes the sequence of embeddings corresponding to all the image patches and text tokens generated by CLIP as input. The module outputs two embeddings incorporating the aggregated image and text information. Instead of directly leveraging these two embeddings for classification, we add a modality-wise attention module to handle the situation where a certain modality (image or text) is irrelevant to the classification task. This is because, in practice, it is common for the image-text pairs to contain noise. $E.g.$, a retailer might upload wrong product images to E-commerce website, or a user might enter random textual comments on social media apps. To handle such situations, we leverage the similar architecture as in \cite{keyless} to learn the importance of each modality to the classification tasks. The sum of the two embeddings weighted by their importances is followed by a MLP head for the classification. To leverage the network for multiple classification tasks, we configure task specific modality-wise attention modules and MLP heads. The complete architecture of CMA-CLIP is shown in Fig. \ref{fig: cma_pipeline}.         

\subsubsection{Sequence-wise Attention}

In our implementation, the sequence-wise attention module is a transformer encoder \cite{NIPS2017_3f5ee243}. Let $X \in \mathbb{R}^{s \times d}$ be the matrix of the sequence of embedding of all the image patches and text tokens generated by CLIP, where $s$ is the length of the sequence and $d$ is the dimension of the embedding. Let $W_K \in \mathbb{R}^{s \times d}$, $W_Q \in \mathbb{R}^{s \times d}$ and $W_V \in \mathbb{R}^{s \times d}$ be the projection matrices which project each embedding in $X$ to key space, query space and value space respectively:
\begin{equation}
K = X W_K , \quad Q =  X W_Q, \quad V =  X W_V
\end{equation}
The embedding matrix $X$ is updated as 
\begin{equation}
\label{eq: att}
\mbox{Attention}(K, Q, V) = \mbox{softmax} (\frac{QK^T}{\sqrt{d_k}})V
\end{equation}
The self-attention block learns a similarity matrix $QK^T$ between each pair of embeddings in $X$. Each embedding in the sequence is then updated as the average of the projected embedding across all the embeddings in the value space weighted by their similarities. This sequence-wise attention module captures the fine-grained relationship between each image patch and text token.

\subsubsection{Modality-wise Attention}

After the image feature and the text feature are generated, they need to be aggregated to form the final feature for the classification tasks. In order to dampen the irrelevant modality, we leverage the keyless attention module proposed in \cite{keyless}. Given an image-text pair, the sequence-wise attention module will output $I'_{[CLS]}$ and $T'_{[CLS]}$ as the global image and text embedding, respectively. The aggregated embedding is the weighted average of $I'_{[CLS]}$ and $T'_{[CLS]}$:

\begin{equation}
\label{eq: aggregation}
c = \lambda I'_{[CLS]} + (1 - \lambda) T'_{[CLS]}
\end{equation}

The weight $\lambda$ is computed by:
\begin{equation}
e_I = w^TI'_{[CLS]}
\end{equation}
\begin{equation}
e_T = w^TT'_{[CLS]}
\end{equation}
\begin{equation}
\lambda = \frac{\exp(e_I)}{\exp(e_I) + \exp(e_T)}
\end{equation}

where $w$ is a learnable parameter vector that is of the same dimension as $I'_{[CLS]}$ and $T'_{[CLS]}$.

\begin{center}
\begin{table}[h]
\begin{tabular}{|| P{0.25\linewidth} P{0.2\linewidth} P{0.06\linewidth} P{0.08\linewidth} P{0.06\linewidth} | P{0.06\linewidth}||}
 \hline
 Method & Data & Color & Pattern & Style & Avg. \\ 
 \hline\hline
Raw CLIP & WIT & 47.3 & 58.0 & 22.1 & 42.5\\
Fine-tuned CLIP & MRWPA & 53.4 & 56.5 & 11.7 & 40.5 \\
CMA-CLIP & MRWPA & \textbf{61.1} & \textbf{76.3} & \textbf{22.9} & \textbf{53.4} \\ 
 \hline
\end{tabular}
\bigskip
 \caption{Recall (\%) at 90\% precision on the MRWPA dataset.}
 \label{tb: pt_results}
\end{table}
\end{center}

\begin{table}[h]
\begin{tabular}{||c c||} 
 \hline
 Method & Food101  \\ 
 \hline\hline
 ViT & 81.8\\
 BERT & 87.2\\
 CLIP & 88.8\\
 MMBT & 92.1 \\
CMA-CLIP& \textbf{93.1}  \\
 \hline
\end{tabular}
\bigskip
 \caption{Accuracies (\%) on the Food101 dataset.}
 \label{tb: food}
\end{table}

\begin{center}
\begin{table}[h]
\begin{tabular}{||c c||} 
 \hline
 Method & Fashion-Gen \\ 
 \hline\hline
 FashionBERT & 85.3\\
 ImageBERT & 80.1\\
 OSCAR & 84.2 \\
 KaleidoBERT & 88.1 \\
CMA-CLIP& \textbf{93.6} \\
 \hline
\end{tabular}
\bigskip
 \caption{Accuracies (\%) on the Fashion-Gen dataset.}
 \label{tb: fashion}
\end{table}
\end{center}

\begin{center}
\begin{table*}[h]
\begin{tabular}{|| m{0.15\linewidth} m{0.15\linewidth} m{0.18\linewidth}  m{0.40\linewidth} ||}
 \hline
 Attribute & Image & Label & Title \\ 
 \hline\hline
 Color &  \includegraphics[scale=0.14]{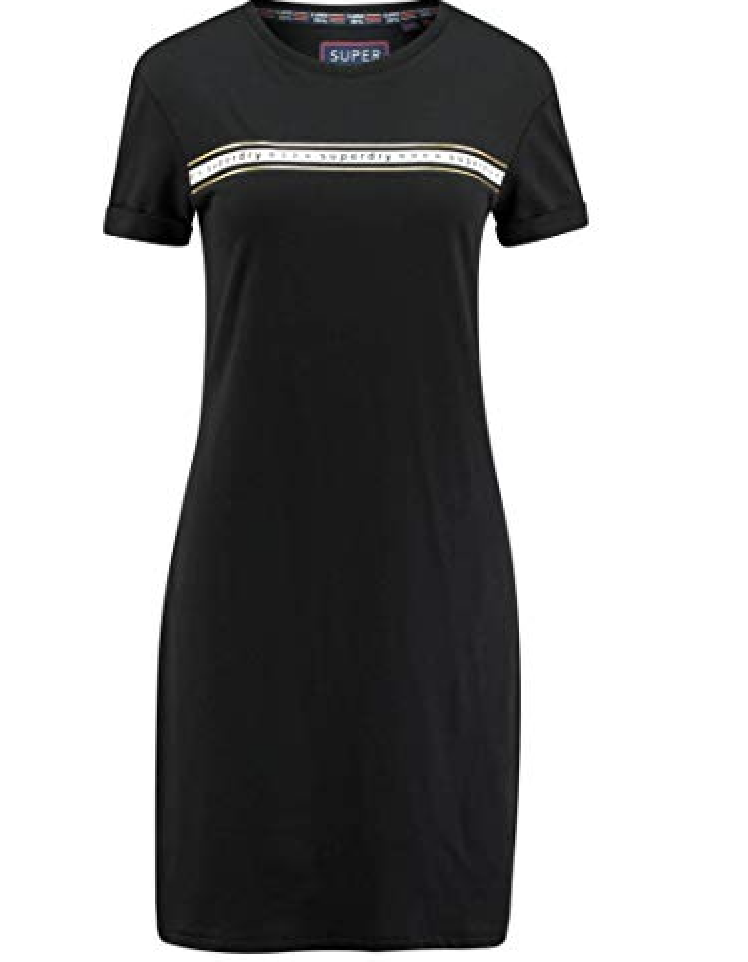} &Black & Portland \textcolor{red}{[Black]} T-Shirt Dress \\ 
 \hline
Pattern &   \includegraphics[scale=0.15]{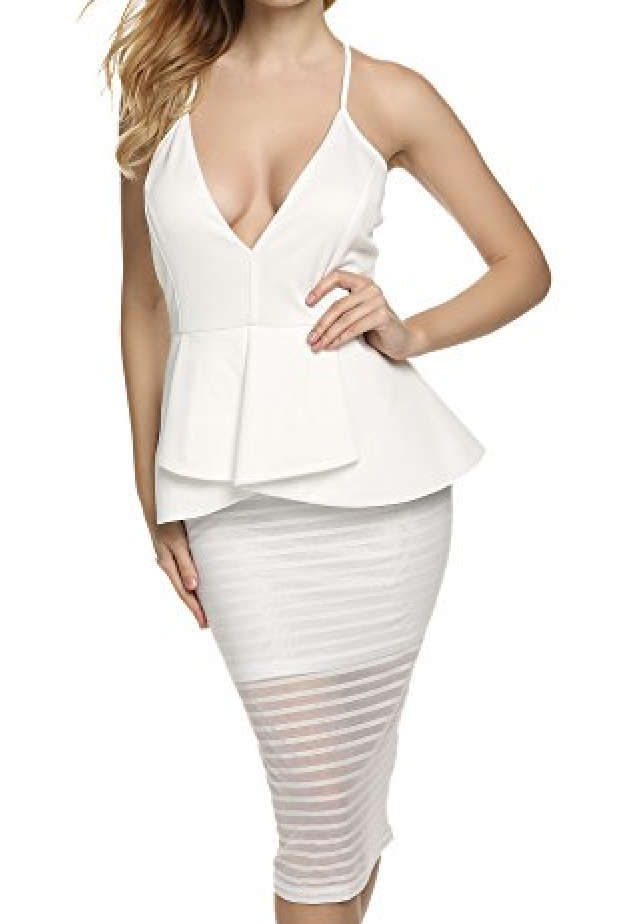} &Plain & Women's Sexy V Neck Crisscross Backless Cocktail Party Bodycon Peplum \textcolor{red}{[Plain]} Dress, White, L \\ 
 \hline
Pattern & \includegraphics[scale=0.16]{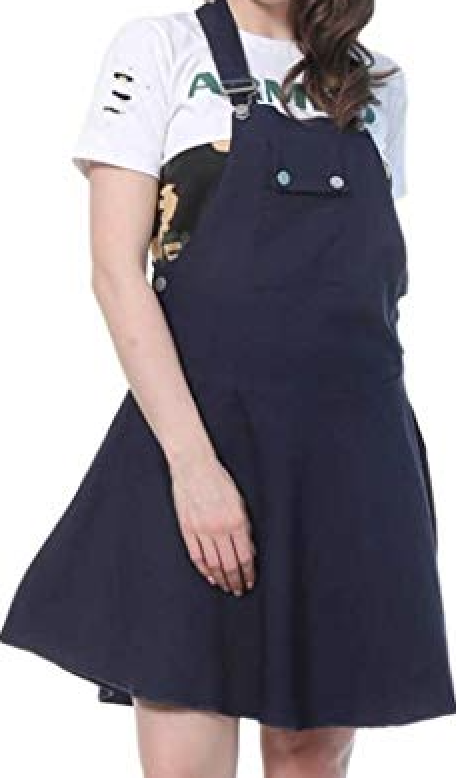} &  Plain & Women's \textcolor{red}{[Plain]} Mini Dungaree\\
 \hline
\end{tabular}
\bigskip
 \caption{Examples where CMA-CLIP is able to give the correct attribute classification while CMA-CLIP w/o the modality-wise attention cannot. Noting that, text tokens which are shown in red are the attribute label keywords that do not exist in the original titles. We add them in and re-do the prediction with both methods to further validate that, without the modality-wise attention, the model is not able to manage noise properly.}
 \label{tb: noisy_examples}
\end{table*}
\end{center}

\subsubsection{MLP Heads for Classification}

For any classification task, an Multi-Layer Perception (MLP) head is added on top of the final feature outputted by the modality-wise attention. The cross entropy is used to compute the loss for this classification task. For multi-task classification, we add task-specific modality-wise attention and MLP for each task separately. This is because the relevance of modality is dependent on the task, hence we need task-specific modality-wise attentions and multiple MLPs as the classification heads. 

\begin{center}
\begin{table}
\begin{tabular}{||c c c c|c||} 
 \hline
 Method & Color & Pattern & Style & Avg. \\ 
 \hline\hline
CMA-CLIP & \textbf{61.1} & \textbf{76.3} & \textbf{22.9} & \textbf{53.4} \\ 
CMA-CLIP w/o $MA$ & 60.0&  67.9 & 14.5 & 47.5\\ 
CMA-CLIP w/o $(MA + SA)$ & 57.3 & 60.3 & 19.8 & 45.8 \\
 \hline
\end{tabular}
\bigskip
 \caption{Ablation study of CMA-CLIP. Recall (\%) at 90\% precision on the MRWPA with Color, Pattern and Style attributes. \textbf{$MA$} denotes modality-wise attention and \textbf{$SA$} denotes to sequence-wise attention.}
 \label{tb: ablation_study}
\end{table}
\end{center}

\begin{figure*}[h]
\centering 
\subfigure[MRWPA: Fashion Women \textcolor{red}{Graphic} Print Round Neck Ringer T-Dress Yellow]{
\includegraphics[width=1.079in]{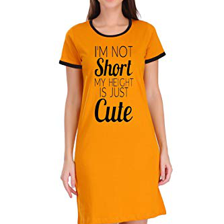}
\includegraphics[width=1.084in]{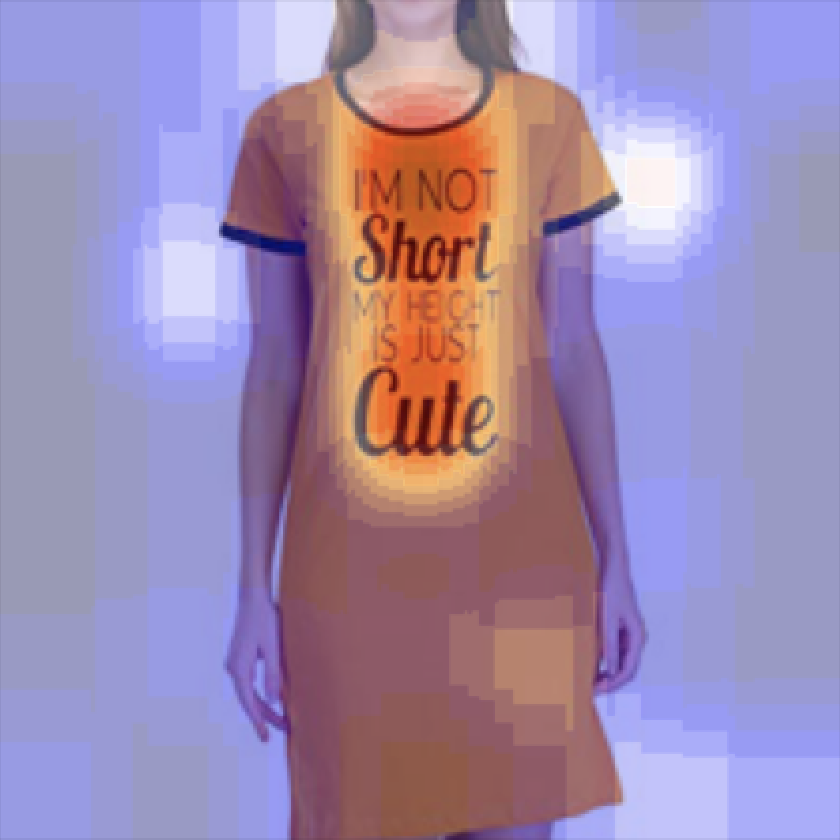}
}
\subfigure[MRWPA: Women's Summer Long Sleeve Casual Loose T-\textcolor{red}{Shirt} Dress]{
\includegraphics[width=1.079in]{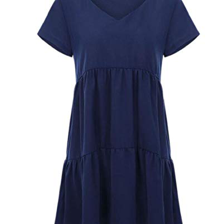}
\includegraphics[width=1.084in]{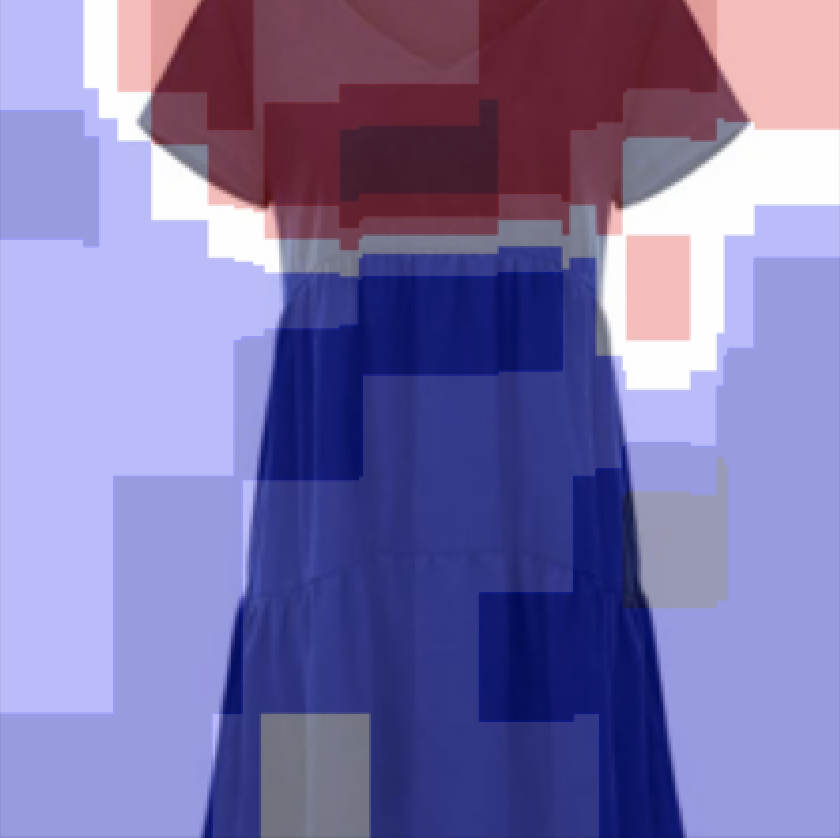}
}
\subfigure[MRWPA: Fashion Women's Oversized Short Sleeves \textcolor{red}{Floral} Print Mid-Long Dress Yellow Size UK 16
]{
\includegraphics[width=1.079in]{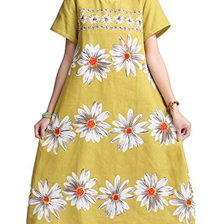}
\includegraphics[width=1.084in]{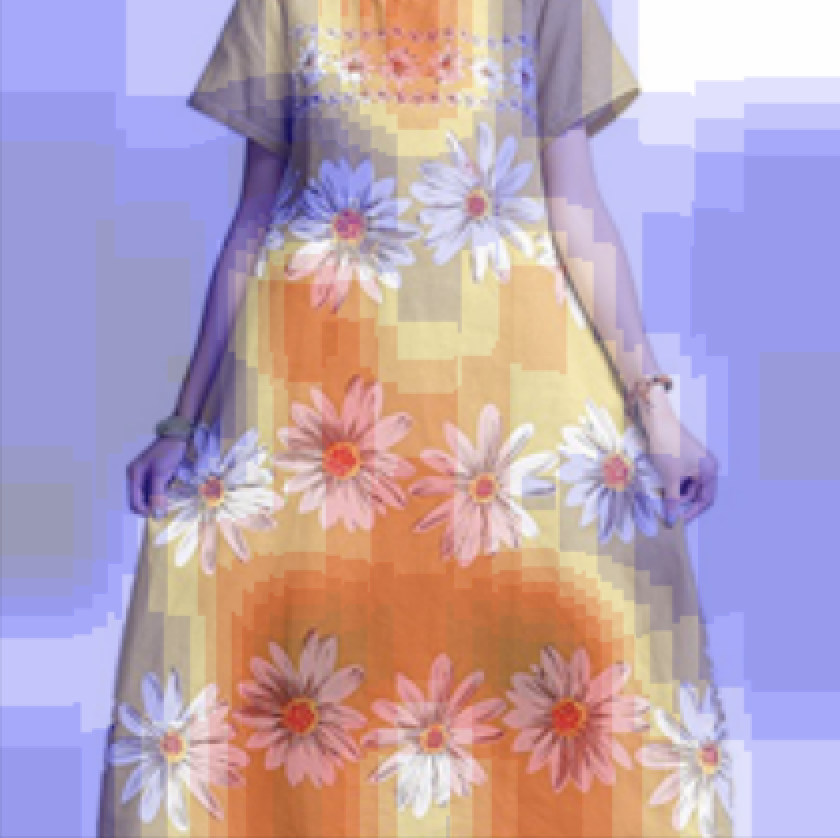}
}
\subfigure[Food101: The Brewer \& The Baker: \textcolor{red}{Lobster} Tail Ale \& Lobster Rolls]{
\includegraphics[width=1.079in]{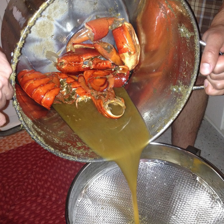}
\includegraphics[width=1.084in]{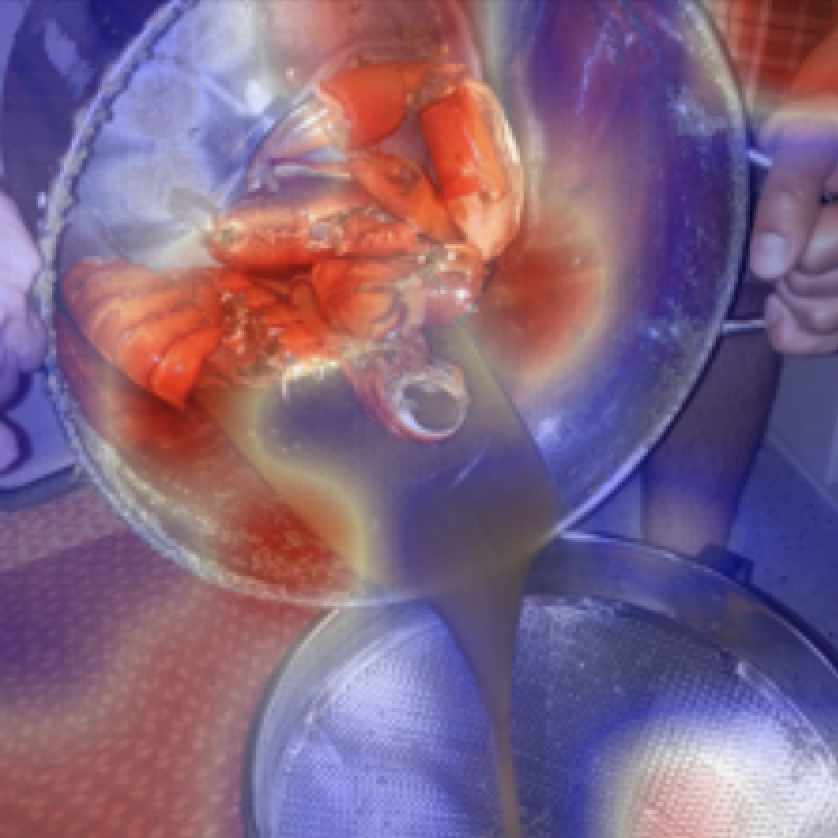}
}
\subfigure[Food101: \textcolor{red}{Scallop} Recipe for Beginners | Pop Sugar Food]{
\includegraphics[width=1.079in]{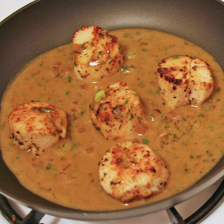}
\includegraphics[width=1.084in]{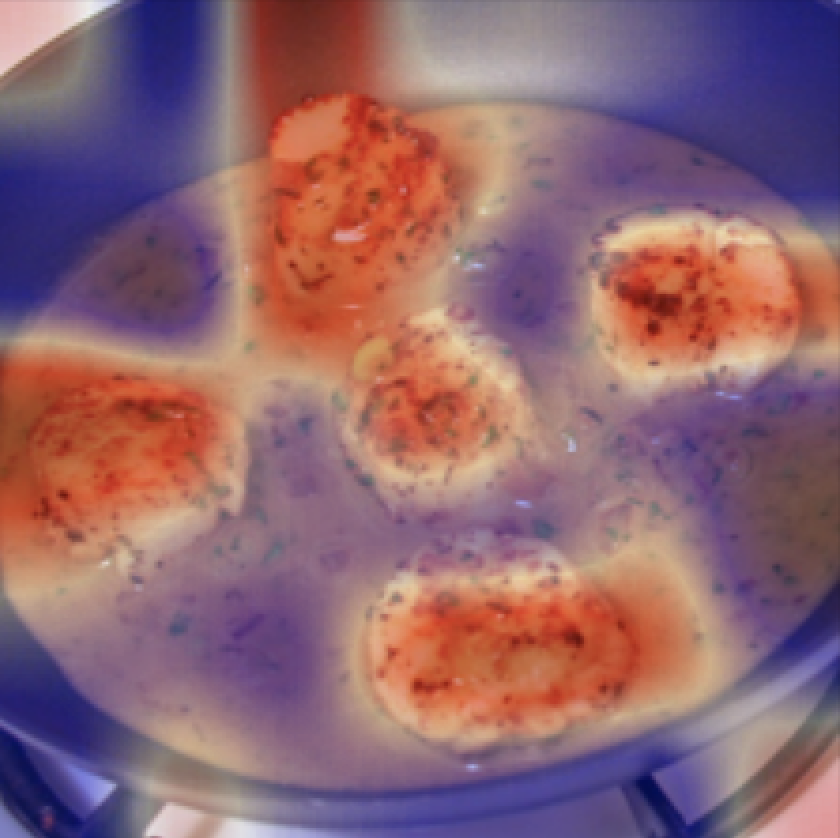}
}
\subfigure[Food101: \textcolor{red}{Shrimp} and Grits - Picture-Perfect Meals \symbol{92}xc2 \symbol{92}xae Picture-Perfect Meals]{
\includegraphics[width=1.079in]{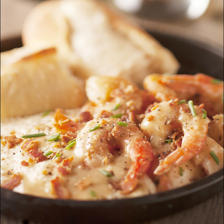}
\includegraphics[width=1.084in]{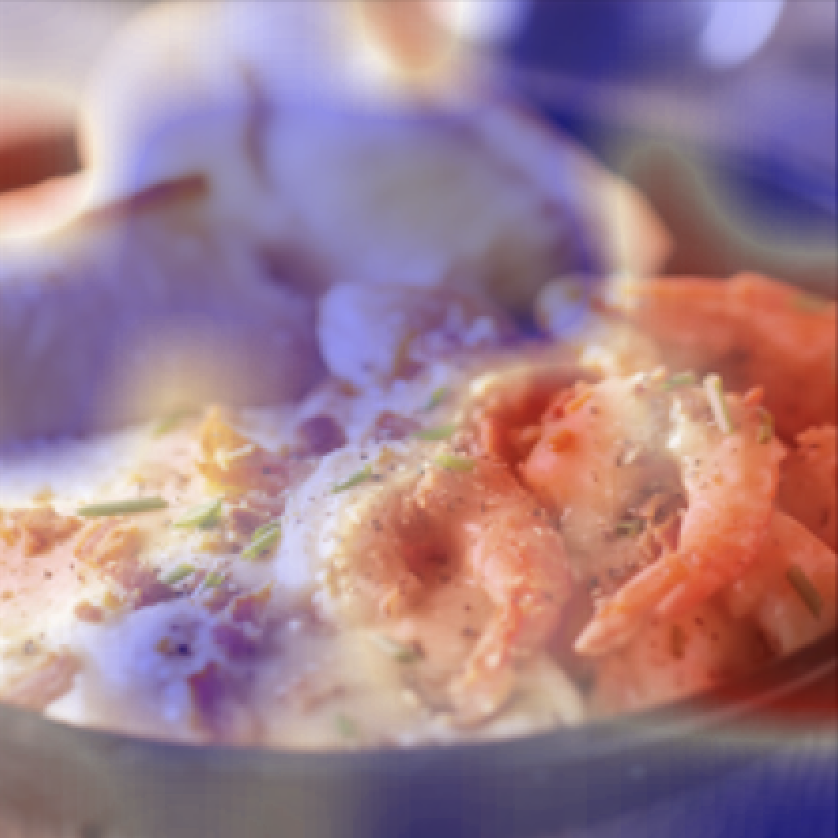}
}
\subfigure[Fashion-Gen: Grained calfskin shoulder \textcolor{red}{bag} in 'plaid' red. Curb chain shoulder strap. Logo stamp gold-tone...]{
\includegraphics[width=1.079in]{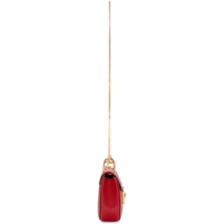}
\includegraphics[width=1.084in]{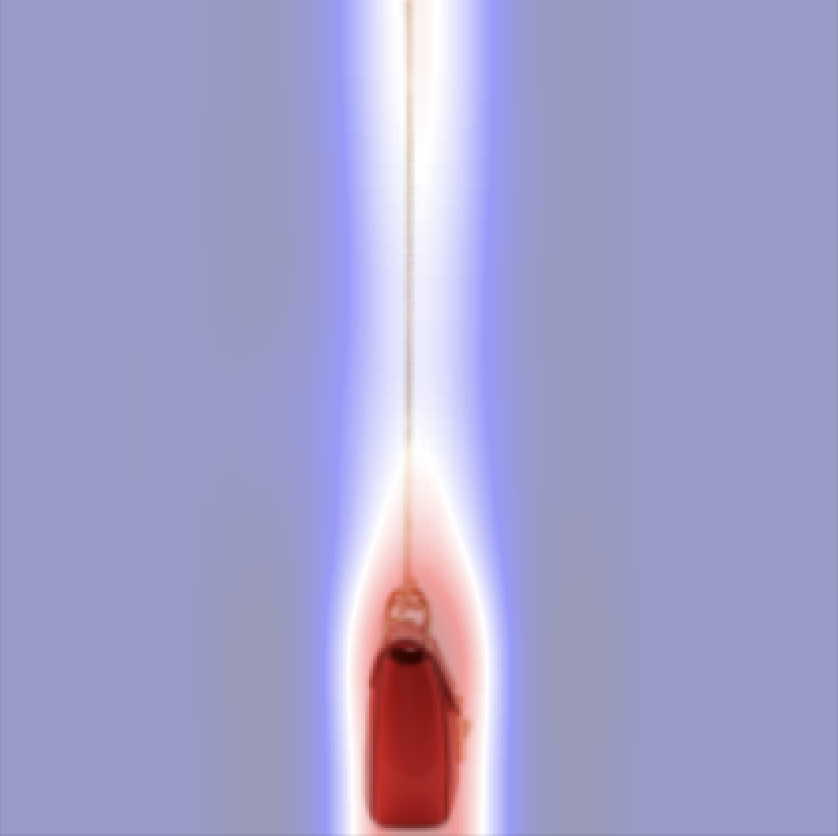}
}
\subfigure[Fashion-Gen: kinny-fit \textcolor{red}{jeans} in mid blue wash. White paint at leg. Low waist. Fading and distressing...]{
\includegraphics[width=1.079in]{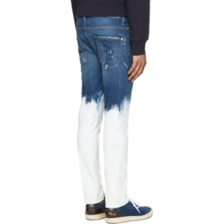}
\includegraphics[width=1.084in]{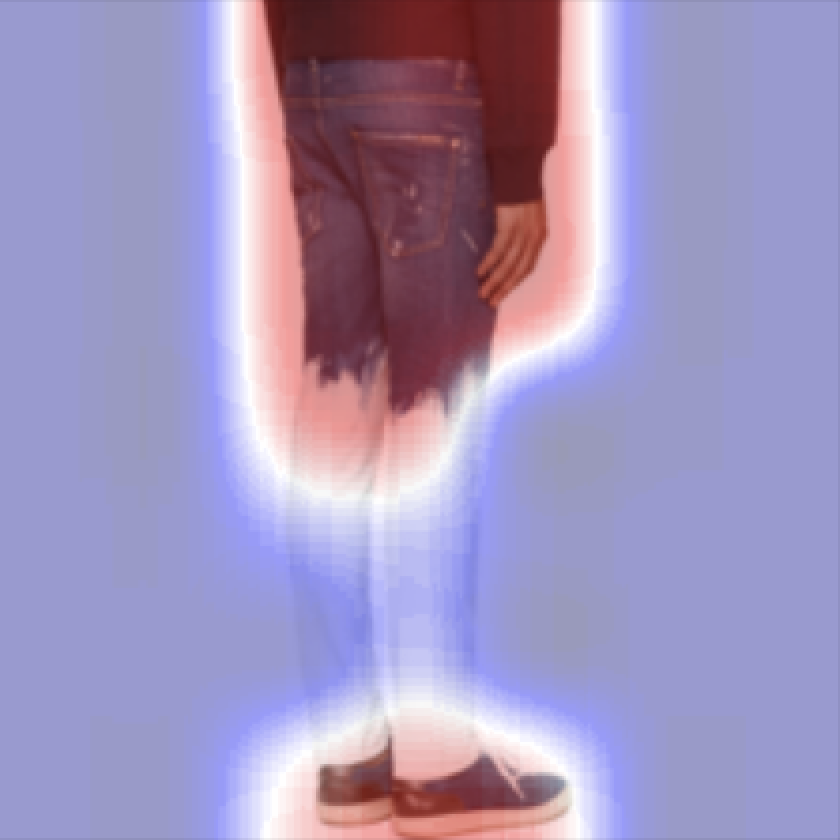}
}
\subfigure[Fashion-Gen: Ankle-high grained leather \textcolor{red}{boots} in black. Pointed toe. Zip closure at heel. Tonal leather...]{
\includegraphics[width=1.079in]{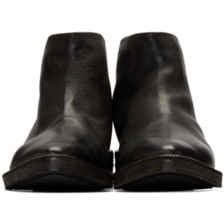}
\includegraphics[width=1.084in]{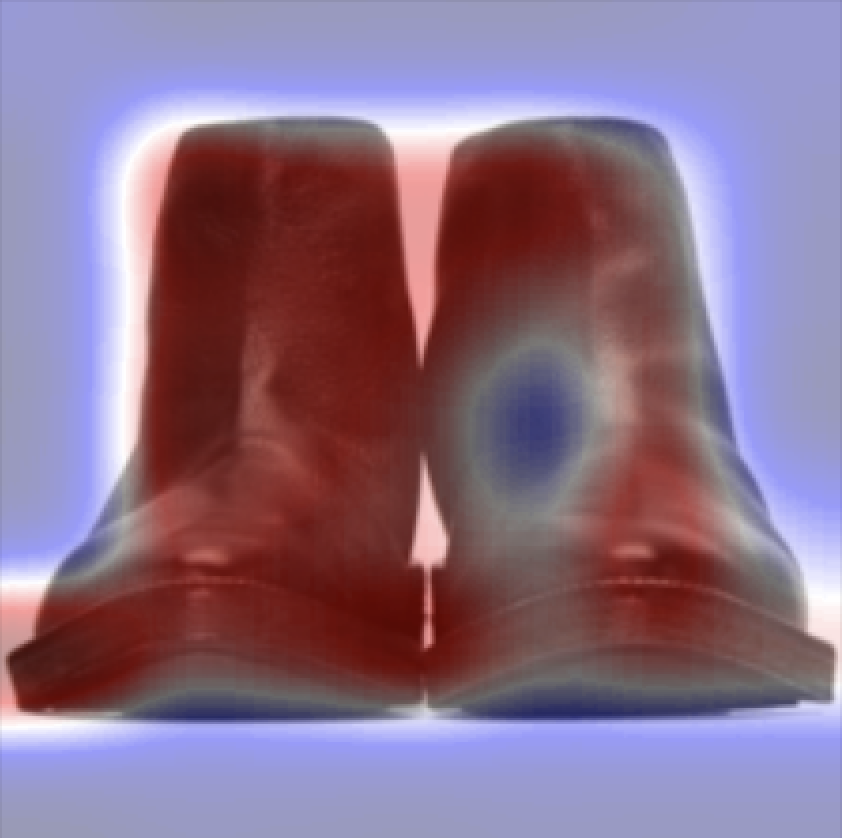}
}
\caption{Visualization examples. Each image is highlighted using the attention map between the image embedding and the embedding of the most relevant text token.}
\label{fig: vis_overall}
\end{figure*}

\section{Experiments}
\subsection{Datasets}
We perform experiments on three datasets, the MRWPA Dataset, the Food101 Dataset \cite{bossard14} and the Fashion-Gen \cite{rostamzadeh2018fashion} Dataset. All three datasets consist of image-text pairs. Data samples from the three datasets are shown in Table \ref{tb: Datasets}.

\subsubsection{The MRWPA Dataset}
This dataset includes the product image and title pairs of dress products from a major retail website. The goal is to classify three dress related product attributes, color, pattern, and style. Color attribute has 17 classes such as black and white, pattern has 12 classes such as graphic and plain, and style has 21 classes such as pencil and a-line. The training data consists of 5.8 Million product image-title pairs. We also prepare 310 and 132 image-title pairs as the validation and test set, which are used for hyper-parameter tuning and performance evaluation respectively.

\subsubsection{The Food101 Dataset}
This dataset contains 101 food categories. The goal is to classify each image-text pair to a food category. We download the preprocessed images and texts from the Kaggle competition\footnote[1]{https://www.kaggle.com/gianmarco96/upmcfood101}. In the processed data, 67971 images are in the training set, and 22715 images are in the testing set. During training, we randomly split 80\% of the data in the training set for training and the rest 20\% data for validation. 

\subsubsection{The Fashion-Gen Dataset}
This dataset contains 293,008 fashion images. Each image is paired with a text describing the image. This dataset contains 48 main categories, such as "DRESSES", "JEANS", "SKIRTS", "SHIRTS", etc.,  and 121 sub-categories, such as "SHORT DRESSES", "LEATHER JACKETS", "MID LENGTH SKIRTS", "T-SHIRTS" and so on. In our experiments, we perform 121 sub-category classification. We use the same data as used in \cite{zhuge2021kaleido} for training and testing. The number of training data is 260480, and the number of testing data is 32528.

\subsection{Implementation and Settings}

\subsubsection{Experiment Settings}
Same as CLIP, the image encoder of CMA-CLIP is a 12-layer 768-width ViT-B/32 \cite{dosovitskiy2021an} with 12 attention heads, and the text encoder of CMA-CLIP is a 12-layer 512-width Transformer with 8 heads used in \cite{NIPS2017_3f5ee243}. The sequence-wise attention transformer is also a 12-layer 512-width model with 8 attention heads. In all the experiments, the batch size is set to 1024, weight decay of Adam is set to $1e-4$, and the learning rate is set to $1e-5$. 

\subsubsection{Training Strategy}
We use the pre-trained weights of CLIP as the initial weights of the image encoder and text encoder in CMA-CLIP. We randomly initialize the weights in the sequence-wise attention module, modality-wise attention module and MLP. As CMA-CLIP contains a mixture of pre-trained weights and randomly initialized weights, instead of training the model end-to-end which may cause under- or over-fitting of certain modules, we adopt a multi-stage training strategy to train CMA-CLIP. The training stages are listed below:
\begin{itemize}
  \item \textbf{Warm-up stage.} In this stage, the weights of the image encoder and the text encoder are frozen. We train the sequence-wise attention, the modality-wise attention and the MLP modules.
  \item \textbf{End-to-end training stage.} In this stage, we unfreeze the weights of the image encoder and the text encoder, and train all the components together. 
  \item \textbf{Tuning stage.} This stage is for multi-task training. The weights of the image encoder, the text encoder and the sequence-wise attention are frozen. We train the modality-wise attentions and MLPs for all the tasks. 
\end{itemize}

\subsubsection{Implementation}
Detailed training process of CMA-CLIP is summarized in Algorithm \ref{alg:cma-clip}. For the MRWPA Dataset, all three stages are trained for 20 epochs. For Food101 and Fashion-Gen Datasets, Warm-up stage is trained for 100 epochs and End-to-end training stage is trained for 300 epochs. Since for the two public datasets, they are both single-task classification so the Tuning stage is not needed. During the Warm-up stage, due to the freeze of the CLIP module, only the check-points of the sequence-wise attention, the modality-wise attention and the MLP modules are updated. During the End-to-end training stage, check-points of all three modules are updated. And during the Tuning stage, only the check-points of the modality-wise attention the MLP modules are updated. At the end of each training stage, the best check-points with the lowest validation accuracy are used for either next stage's fine-tuning or inference.

\subsection{Experimental Results}
\subsubsection{The MRWPA Dataset}
We compare CMA-CLIP with the zero-shot performance of raw CLIP and fine-tuned CLIP (fine-tuned using image-title pairs in MRWPA dataset) in terms of the recall at 90\% precision for the color, pattern, and style attributes. The results are included in Table \ref{tb: pt_results}. We observe that CMA-CLIP consistently outperforms both raw CLIP and fine-tuned CLIP by a large margin across all three attributes. 

\subsubsection{The Food101 Dataset}
On the Food101 dateset, we compare CMA-CLIP with two single-modality baseline methods including BERT \cite{devlin2018bert} and ViT \cite{dosovitskiy2021an}, and two multi-modality baseline methods including raw CLIP \cite{radford2021learning} with same ViT-B/32 and MMBT \cite{kiela2019supervised}. Results are included in Table \ref{tb: food}. CMA-CLIP achieves the best accuracy of 93.1\%, which improves 1\% over the a current strong baseline method MMBT. Using only image features achieves 81.8\% by ViT, and using only text features achieves 87.2\% by BERT.

\subsubsection{The Fashion-Gen Dataset}
On the Fashion-Gen dataset, we compare CMA-CLIP with multiple SOTA methods including FashionBERT \cite{gao2020fashionbert}, ImageBERT \cite{qi2020imagebert}, OSCAR \cite{li2020oscar} and KaleidoBERT \cite{zhuge2021kaleido}. CMA-CLIP achieves the highest accuracy of 93.6\%, which improves over KaleidoBERT \cite{zhuge2021kaleido}, the previous SOTA method, by 5.5\%. 

\subsection{Ablation Study}
We conduct systematic ablation study to validate the effectiveness of modality-wise attention module and sequence-wise attention module by removing them sequentially and comparing the performance with CMA-CLIP. Detailed results are shown in Table \ref{tb: ablation_study}.

On MRWPA, the average recall at 90\% precision across the 3 attributes drops from 53.4\% to 47.5\% after removing the modality-wise attention module. This is because the proportions of titles that contain tokens related to color, pattern, and style are 67\%, 25\% and 15\% respectively. When a title does not contain any tokens related to an attribute, it becomes irrelevant for the classification of that attribute. The performance drop indicates that the modality-wise attention module significantly improves CMA-CLIP's robustness against noisy inputs. 

To illustrate our model's robustness to input noise, in Table \ref{tb: noisy_examples} we randomly pick some product image-title examples that CMA-CLIP is able to give correct classification whereas CMA-CLIP without modality-wise attention module cannot. We can clearly observe that in those examples, the product titles do not contain any tokens related to the attribute labels. Furthermore, for these examples, we complete the titles by adding the label related keywords and re-test them. This time, both methods can provide correct classification results which further proves that the modality-wise attention has the ability of filtering out irrelevant information (text without label information is considered as noise). We are not able to select similar examples in the Food101 and Fashion-Gen datasets, because in these two public datasets, there are no images or text that are irrelevant to the classification task.     

The average recall drops from 47.5\% to 45.8\% after further removing the sequence-wise attention module. The sequence-wise attention module enhances the context-awareness of the image and text embedding by capturing the fine-grained correlation among image patches and text tokens, and the resulting embedding is expected to yield better results for classifications. The performance drop supports this conclusion.

We also visualize the result of sequence-wise attention for MRWPA, Food101 and Fashion-Gen datasets in Figure \ref{fig: vis_overall}. For each text input, we locate the token that is related to the classification task, and visualize the image patches that are most correlated to it by checking the inner product between the query embedding of the text token and the key embeddings of the image patches. In Figure \ref{fig: vis_overall}, red regions are where the correlation is high. We observe that the sequence-wise attention is able to identify the highly correlated image patches and text tokens across all three datasets.

\section{Conclusion}
In this paper, we propose the CMA-CLIP, which unifies two types of cross-modality attentions: sequence-wise attention, a transformer based attention module that captures the fine-grained relationship between image patches and text tokens, and modality-wise attention, which learns the importance of image and text modalities in order to filter out the irrelevant modality for the classification task. We also design task specific modality-wise attentions and MLPs so that we can leverage a unified network for multi-task classifications. We evaluate our method on the MRWPA Dataset, the Food101 dataset and the Fashion-Gen dataset. CMA-CLIP outperforms the pre-trained and fine-tuned CLIP by an average of 11.9\% in recall at the same level of precision on the MRWPA Dataset for the classifications for color, pattern, and style attributes. It also surpasses the state-of-the-art method on the Fashion-Gen Dataset by 5.5\% in accuracy and achieves competitive performance on the Food101 Dataset. For the future work, we are interested in training CMA-CLIP with other datasets to enable the contrastive loss, improving CMA-CLIP's robustness against noisy labels, and also, exploring semi-supervised learning methods so that unlabeled image-text pairs can be leveraged in the training process to improve model generalizability.
\bibliographystyle{ACM-Reference-Format}
\bibliography{references}

\end{document}